# Twitter Sentiment Analysis System

Shaunak Joshi
Department of Information Technology
Vishwakarma Institute of Technology
Pune, Maharashtra, India

Deepali Deshpande
Department of Information Technology
Vishwakarma Institute of Technology
Pune, Maharashtra, India

## ABSTRACT
Social media is increasingly used by humans to express their feelings and opinions in the form of short text messages. Detecting sentiments in text has a wide range of applications including identifying anxiety or depression of individuals and measuring well-being or mood of a community. Sentiments can be expressed in many ways that can be seen such as facial expression and gestures, speech and by written text. Sentiment Analysis in text documents is essentially a content – based classification problem involving concepts from the domains of Natural Language Processing as well as Machine Learning. In this paper, sentiment recognition based on textual data and the techniques used in sentiment analysis are discussed.

## 1. INTRODUCTION
What do you do when you want to express yourself or reach out to a large audience? We log on to one of our favorite social media services. Social Media has taken over in today's world, most of the methods we use to connect and communicate are using social networks, and Twitter is one of the major places where we express our sentiments about a specific topic or a concept.

Twitter serves as a mean for individuals to express their thoughts or feelings about different subjects. [1] These emotions are used in various analytics for better understanding of humans. [2] In this paper, we have attempted to conduct sentiment analysis on "tweets" using different machine learning algorithms. We attempt to classify the polarity of the tweet where it is either positive or negative. If the tweet has both positive and negative elements, the more dominant sentiment should be picked as the final label.

## 2. PROBLEM STATEMENT & APPLICABILITY
Since the Advent of the Internet, humans have used it as a communication tool in the form of mostly text messages and nowadays video and audio streams and as we increase our dependence on technology it becomes increasingly important to better gauge human sentiments expressed with the help of technology. However, in this textual communication data, we lose the access to sentiments or the emotions conveyed behind a sentence, as we often use our hands and facial expressions to express our intent behind the statement. From this textual data, we can gain insights into the individual. Insights which can be used for multiple different uses such as content recommendation based on current mood, market segmentation analysis and psychological analysis in humans. [3]

In this project, we have attempted to classify human sentiment into two categories namely positive and negative. Which helps us better understand human thinking and gives us an insight which can be used in a variety of ways as stated above.

## 3. PROPOSED METHODOLOGY
In this paper we classify sentiments with the help of machine learning and natural language processing (NLP) algorithms, we use the datasets from Kaggle which was crawled from the internet and labeled positive/negative. The data provided comes with emoticons (emoji), usernames and hashtags which are required to be processed (so as to be readable) and converted into a standard form. We also need to extract useful features from the text such unigrams and bigrams which is a form of representation of the "tweet". We use various machine learning algorithms based on NLP (Natural Language Processing) to conduct sentiment analysis using the extracted features. Finally, we report our experimental results and findings at the end.

### 3.1 Data Description
The data given from the dataset is in the form of comma-separated values files with "tweets" and their corresponding sentiments. The training dataset is a csv (comma separated value) file of type tweet_id, sentiment, tweet where the tweet_id is a unique integer identifying the tweet, sentiment is either 1 (positive) or 0 (negative), and tweet is the tweet enclosed in "<inverted brackets>". Similarly, the test dataset is a csv file of type tweet_id, tweet respectively. The dataset is a mixture of words, emoticons, symbols, URLs and references to people as seen usually on twitter. Words and emoticons contribute to predicting the sentiment, but URLs and references to people don't. Therefore, URLs and references are being ignored. The words are also a mixture of misspelled words / incorrect, extra punctuations, and words with many repeated letters. The "tweets", therefore, must be preprocessed to standardize the dataset. The provided training and test dataset have 800000 and 200000 tweets respectively. Preliminary statistical analysis of the contents of datasets, after preprocessing as described in section 3.1, is shown in tables 1 and 2.

**Table 1: Statistics of Preprocessed Train Dataset**

|  | Total | Unique | Average | Maximum | Positive | Negative |
|---|---|---|---|---|---|---|
| Tweets | 800000 | N/A | N/A | N/A | 400312 | 399688 |
| User Mentions | 393392 | N/A | 0.4917 | 12 | N/A | N/A |
| Emoticons | 6797 | N/A | 0.0085 | 5 | 5807 | 990 |
| URLs | 38698 | N/A | 0.0484 | 5 | N/A | N/A |
| Unigrams | 9823554 | 181232 | 12.279 | 40 | N/A | N/A |
| Bigrams | 9025707 | 1954953 | 11.28 | N/A | N/A | N/A |

**Table 2: Statistics of Preprocessed Test Dataset**

|  | Total | Unique | Average | Maximum | Positive | Negative |
|---|---|---|---|---|---|---|
| Tweets | 200000 | N/A | N/A | N/A | N/A | N/A |
| User Mentions | 97887 | N/A | 0.4894 | 11 | N/A | N/A |
| Emoticons | 1700 | N/A | 0.0085 | 10 | 1472 | 228 |
| URLs | 9553 | N/A | 0.0478 | 5 | N/A | N/A |
| Unigrams | 2457216 | 78282 | 12.286 | 36 | N/A | N/A |
| Bigrams | 2257751 | 686530 | 11.29 | N/A | N/A | N/A |

## 3.2 Preprocessing

Raw tweets scraped from twitter generally result in a noisy and obscure dataset. This is due to the casual and ingenious nature of people's usage of social media. Tweets have certain special characteristics such as retweets, emoticons, user mentions, etc. which should be suitably extracted. Therefore, raw twitter data must be normalized to create a dataset which can be easily learned by various classifiers. We have applied an extensive number of pre-processing steps to standardize the dataset and reduce its size. We first do some general pre-processing on tweets which is as follows:

- Convert the tweet characters to lowercase alphabet.
- Replace 2 or more dots (.) with space
- Strip spaces and quotes (" and ') from the ends of tweet.
- Replace 2 or more spaces with a single space.

We handle special twitter features as follows:

### 3.2.1 Uniform Resource Locator (URL)

Users often share hyperlinks to other webpages in their tweets. Any particular given URL is not important for text classification as it would lead to very sparse features and incorrect classification. Therefore, we replace all the URLs in tweets with the word URL. The regular expression used to match URLs is ((www\.[\S]+)|(https?://[\S]+)).

### 3.2.2 User Mention

Every twitter user has a handle associated with them. Users often mention other users in their tweets by @handle. We replace all user mentions with the word USER_MENTION. The regular expression (regex) used to match user mention is @[\S]+. 2.

### 3.2.2 Emoticon

Users often use several different emoticons in their tweet to convey different emotions. It is impossible to exhaustively match all the different emoticons used on social media as the number is ever increasing. However, we match some common emoticons which are used very frequently. We replace the matched emoticons with either EMO_POS or EMO_NEG depending on whether it is conveying a positive or a negative emotion. A list of all emoticons matched by our method is given in table 3.

**Table 3: List of Emoticons matched by our method**

| Emoticon(s) | Type | Regex | Replacement |
|---|---|---|---|
| :), : ), :-), (:, (:, (-:, :') | Smile | (:\s?\)\|:-\)\|\(\s?:\|\(-:\|:\'\)) | EMO_POS |
| :D, : D, :-D, xD, x-D, XD, X-D | Laugh | (:\s?D\|:-D\|x-?D\|X-?D) | EMO_POS |
| ;-), ;), ;-D, ;D, (;, (-; | Wink | (:\s?\(\|:-\(\|\)\s?:\|\)-:) | EMO_POS |
| <3, :* | Love | (<3\|:\*) | EMO_POS |
| :-(, : (, :(, ):, )-: | Sad | (:\s?\(\|:-\(\|\)\s?:\|\)-:) | EMO_NEG |
| :,(, :'(, :"( | Cry | (:,\(\|:\'\(\|:"\() | EMO_NEG |

### 3.2.3 Hashtag
Hashtags are un-spaced phrases prefixed by the hash symbol (#) which is frequently used by users to mention a trending topic on twitter. We replace all the hashtags with the words with the hash symbol. For example, #hello is replaced by hello. The regular expression used to match hashtags is #(\S+).

### 3.2.4 Retweet
Retweets are tweets which have already been sent by someone else and are shared by other users. Retweets begin with the letters RT. We remove RT from the tweets as it is not an important feature for text classification. The regular expression used to match retweets is \brt\b.

After applying tweet level pre-processing, we processed individual words of tweets as follows:

- Strip any punctuation ['"?!,.():;] from the word.
- Convert 2 or more letter repetitions to 2 letters. Some people send tweets like *I am sooooo happpppy* adding multiple characters to emphasize on certain words. This is done to handle such tweets by converting them to *I am soo happy*.
- Remove - and '. This is done to handle words like t-shirt and their's by converting them to the more general form tshirt and theirs.
- Check if the word is valid and accept it only if it is. We define a valid word as a word which begins with an alphabet with successive characters being alphabets, numbers or one of dot (.) and underscore(_).

Some example tweets from the training dataset and their normalized versions are shown in table 4.

**Table 4: Example Tweets from the Dataset and their Normalized Version**

| Raw | misses Swimming Class. http://plurk.com/p/12nt0b |
|---|---|
| Normalized | misses swimming class URL |
| Raw | @98PXYRochester HEYYYYYYYYY!! its Fer from Chile again |
| Normalized | USER_MENTION heyy its fer from chile again |
| Raw | Sometimes, You gotta hate #Windows updates. |
| Normalized | sometimes you gotta hate windows updates |
| Raw | @Santiago_Steph hii come talk to me i got candy :) |
| Normalized | USER_MENTION hii come talk to me i got candy EMO_POS |
| Normalized | @bolly47 oh no :'( r.i.p. your bella |
| Raw | USER_MENTION oh no EMO_NEG r.i.p your bella |

## 3.3 Feature Extraction
We extract two types of features from our dataset, namely unigrams and bigrams. We create a frequency distribution of the unigrams and bigrams present in the dataset and choose top N unigrams and bigrams for our analysis.

### 3.3.1 Unigrams
Probably the simplest and the most commonly used features for text classification is the presence of single words or tokens in the text. We extract single words from the training dataset and create a frequency distribution of these words. A total of 181232 unique words are extracted from the dataset. Out of these words, most of the words at end of frequency spectrum are noise and occur very few times to influence classification. We, therefore, only use top N words from these to create our vocabulary where N is 15000 for sparse vector classification.

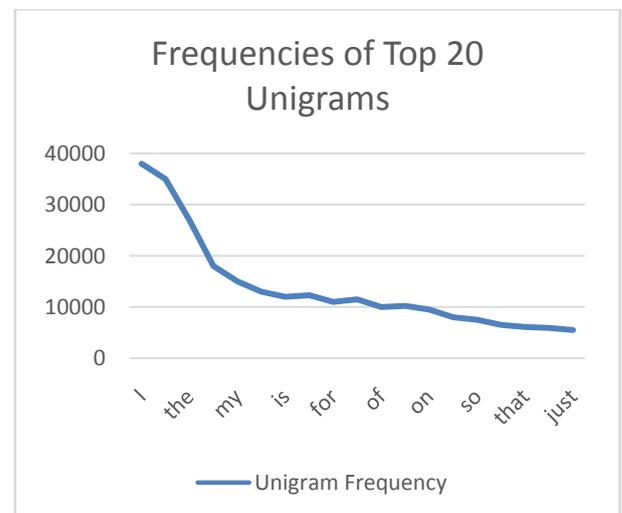

**Figure 1: Statistics of Unigram Occurrence Frequency**

### 3.3.2 Bigrams

Bigrams are word pairs in the dataset which occur in succession in the corpus. These features are a good way to model negation in natural language like the phrase– *This is not good*. A total of 1954953 unique bigrams were extracted from the dataset. Out of these, most of the bigrams at end of frequency spectrum are noise and occur very few times to influence classification. We therefore use only top 10000 bigrams from these to create our vocabulary.

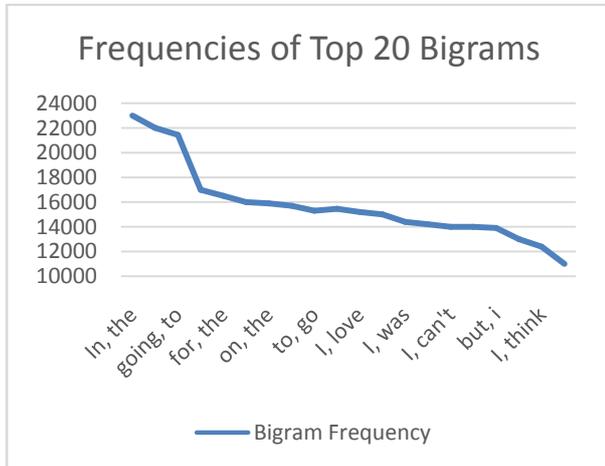

**Figure 2: Statistics of Bigram Occurrence Frequency**

## 4. MODEL

We will be using a model consisting of both test and training dataset model using various algorithms as due to the modular nature of the program we can add and remove the algorithms with ease. Let's understand the workflow of our system with the help of above diagram. First, we have split the data into training and test set. We also keep separate positive and negative pre-labelled datasets for training the model and checking their generalization classification in test set. After this the training data is fed to some machine learning algorithm like *Naive Bayes*, *Maximum Entropy* or *SVM* [4] which learns to make predictions. To evaluate our system, we use Baseline Classification which is our evaluation metric in which test data is fed to the learned algorithm which in return generates recommended prediction ratings of words. With help of pre-classified golden set and evaluation metric we check the accuracy of our model.

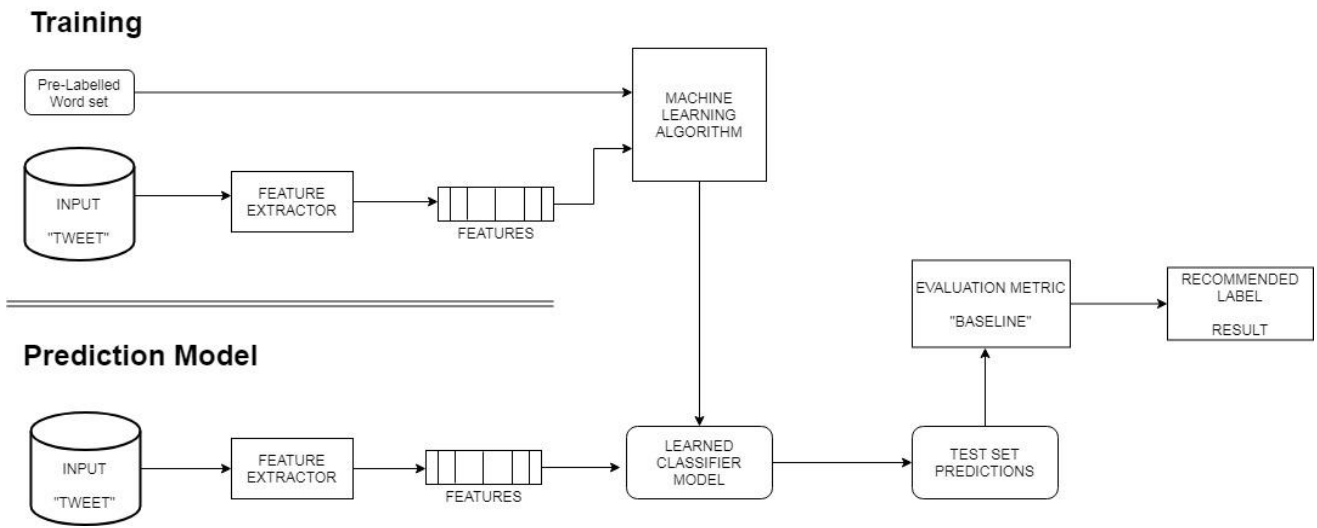

**Fig 3: Machine Learning Model for Twitter System Analysis System**

### 4.1 Algorithms

#### 4.1.1 Baseline (Evaluation Metric)

For a baseline, we use a simple positive and negative word counting method to assign sentiment to a given tweet. We use the Opinion Dataset of positive and negative words to classify tweets. In cases when the numbers of positive and negative words are equal, we assign positive sentiment.

#### 4.1.2 Naïve Bayes

Naive Bayes classifiers are a family of simple "probabilistic classifiers "based on applying Bayes' theorem with strong (naive) independence assumptions between the features." [5]

Naive Bayes is a simple model which can be used for text classification. In this model, the class $\hat{c}$ is assigned to a tweet t, where

$$\hat{c} = P(c|t) argmax_c$$

$$P(c|t) \propto P(c) \prod_{i=1}^{n} P(f_i|c)$$

In the formula above, $f_i$ represents the $i$-th feature of total $n$ features. $P(c|t)$ and $P(f_i|c)$ can be obtained through maximum likelihood estimates. We used `MultinomialNB` from `sklearn.naive_bayes` package of *scikit-learn* for Naive Bayes classification. We used Laplace smoothed version of Naive Bayes with the smoothing parameter α set to its default value of 1. We used sparse vector representation for classification and ran experiments using both presence and frequency feature types. We found that presence features outperform frequency features because Naive Bayes is essentially built to work better on integer features rather than floats. We also observed that addition of bigram features improves the accuracy.

### *4.1.3 Maximum Entropy*

Maximum Entropy Classifier model is based on the Principle of Maximum Entropy. The main idea behind it is to choose the most uniform probabilistic model that maximizes the entropy, with given constraints. [6] Unlike Naive Bayes, it does not assume that features are conditionally independent of each other. So, we can add features like bigrams without worrying about feature overlap. The model is represented by

$$P_{ME}(c|d,\lambda) = P \frac{\exp[\sum_i \lambda_i f_i(c,d)]}{\sum_{c'} \exp[\sum_i \lambda_i f_i(c,d)]}$$

Here, $c$ is the class, $d$ is the tweet and $\lambda$ is the weight vector. The weight vector is found by numerical optimization of the lambdas to maximize the conditional probability.

The nltk library provides several text analysis tools. We use the MaxentClassifier to perform sentiment analysis on the given tweets. Unigrams, bigrams and a combination of both were given as input features to the classifier. The Improved Iterative Scaling algorithm for training provided better results than Generalized Iterative Scaling

## 5. EVALUATION METRICS

For evaluation metrics, we use the baseline algorithm which uses simple positive and negative word counting method to assign sentiment to a given tweet. We use the Golden Dataset of positive and negative words to classify tweets. In cases when the magnitude of positive and negative words is equal, we assign positive sentiment. A baseline is a method that uses heuristics, simple summary statistics, randomness, or machine learning to create predictions for a dataset. We can use these predictions to measure the baseline performance (e.g. accuracy) this metric will then become what we compare any other machine learning algorithm against.

## 6. CONCLUSION

We tried to build a sentiment analysis system by studying and implementing algorithms of machine learning. We implemented Naive Bayes and Maximum Entropy algorithms. Baseline model performed the worst with no doubt as it had least number of features. The modular system we've built can easy be scaled for new algorithms be it in Machine Learning, Deep learning or Natural Language Processing. Sentiment analysis system is an active field of research and we can still further improve our system by working more on the algorithms, trying out different things in preprocessing and checking which ones get the best precision metrics.

## 6.1 Future Work

### *6.1.1 Handling Emotion Ranges*

We can improve and train our models to handle a range of sentiments. Tweets don't always have positive or negative sentiment. At times they may have no sentiment i.e. neutral. Sentiment can also have gradations like the sentence, *This is good*, is positive but the sentence, *This is extraordinary*. Is somewhat more positive than the first. We can therefore classify the sentiment in ranges; say from -2 to +2.

### *6.1.2 Using symbols*

During our pre-processing, we discard most of the symbols like commas, full-stops, and exclamation mark. These symbols may be helpful in assigning sentiment to a sentence.

## 7. ACKNOWLEDGMENTS

We wish to express our thanks towards Prof. Deshpande, Dr. Ghadekar and our families as without their constant support and guidance this project would have not been possible.